\documentclass[conference]{IEEEtran}
\IEEEoverridecommandlockouts
\usepackage{cite}
\usepackage{amsmath,amssymb,amsfonts}
\usepackage{algorithmic}
\usepackage{graphicx}
\usepackage{textcomp}
\usepackage{xcolor}
\usepackage{caption}
\usepackage{subcaption}
\usepackage{bbold}
\def\BibTeX{{\rm B\kern-.05em{\sc i\kern-.025em b}\kern-.08em
    T\kern-.1667em\lower.7ex\hbox{E}\kern-.125emX}}
\begin{document}

\newcommand{\red}[1]{\textcolor{red}{[#1]}}
\newcommand{\blue}[1]{\textcolor{blue}{[#1]}}
\newcommand{\teal}[1]{\textcolor{teal}{#1}} 

\newcommand{\fref}[1]{Fig.~\ref{#1}}
\newcommand{\tref}[1]{Table~\ref{#1}}
\newcommand{\sref}[1]{Section~\ref{#1}}

\title{Learning Deep Representations via Contrastive Learning for Instance Retrieval\\
}

\author{\IEEEauthorblockN{Tao Wu and Tie Luo$^*$\thanks{* Corresponding author.}}
\IEEEauthorblockA{{Department of Computer Science} \\
{Missouri University of Science and Technology}\\
Rolla, MO 65409, USA \\
\{wuta, tluo\}@mst.edu}
\and
\IEEEauthorblockN{Donald C. Wunsch II}
\IEEEauthorblockA{{Department of Electrical \& Computer Engineering} \\
{Missouri University of Science and Technology}\\
Rolla, MO 65409, USA \\
dwunsch@mst.edu}
}

\maketitle

\begin{abstract}
Instance-level Image Retrieval (IIR), or simply Instance Retrieval, deals with the problem of finding all the images within an dataset that contain a query instance (e.g. an object). This paper makes the first attempt that tackles this problem using {\em instance-discrimination} based {\em contrastive learning} (CL). While CL has shown impressive performance for many computer vision tasks, the similar success has never been found in the field of IIR. In this work, we approach this problem by exploring the capability of deriving discriminative representations from pre-trained and fine-tuned CL models. To begin with, we investigate the efficacy of transfer learning in IIR, by comparing off-the-shelf features learned by a pre-trained deep neural network (DNN) classifier with features learned by a CL model. The findings inspired us to propose a new training strategy that optimizes CL towards learning IIR-oriented features, by using an Average Precision (AP) loss together with a fine-tuning method to learn contrastive feature representations that are tailored to IIR. Our empirical evaluation demonstrates significant performance enhancement over the off-the-shelf features learned from a pre-trained DNN classifier on the challenging Oxford and Paris datasets.
\end{abstract}


\section{Introduction}

Instance-level Image Retrieval (IIR) aims to find all the images in a large database that contain a given query content. This technique has a wide range of potential applications such as online product search \cite{liu2016deepfashion} and automatic organization of personal photos \cite{guy2018care}. Since candidate images and the query object are subject to various conditions such as different backgrounds, imaging distances, view angles, illuminations, and weather conditions, successful IIR is challenging and it is crucial to extract discriminative and compact representations for both images and the query object. To obtain effective feature representations for IIR, mainstream techniques have evolved from {\em feature engineering} to {\em deep learning}. The former category was dominated by handcrafted techniques such as SIFT \cite{lowe2004distinctive} and SURF \cite{bay2008speeded} descriptors, while the latter category features the powerful workhorse of deep neural networks (DNNs). DNNs are able to automatically learn more effective representations with multiple levels of abstraction from data, and have set the state of the art in a large variety of classical computer vision tasks including image classification \cite{he2016deep}, object detection \cite{ren2015faster}, semantic segmentation \cite{minaee2021image}, and image retrieval \cite{gordo2017end}.

Prior work based on deep learning often uses the feature maps in the intermediate layers of DNN classifiers pretrained on ImageNet as the descriptor for an image. Following this, feature aggregation methods such as BoW\cite{mohedano2016bags} and R-MAC\cite{tolias2015particular} are used to aggregate these off-the-shelf features into a compact representation. The limitation of such methods is a {\em domain-shift} problem \cite{azizpour2015factors}, which means that a model trained for one task, e.g., classification, does not necessarily extract features suitable for another task, e.g., image retrieval. In particular, the features extract by a DNN trained for classification can be sub-optimal for IIR since IIR requires more fine-grained and more discriminative representations, which typically cannot be satisfied by off-the-shelf features learned for classification.

Recently, contrastive learning (CL) emerged as a powerful approach to self-supervised learning when working with unlabeled data. Several CL studies have reported impressive performance improvement over transfer learning on classification and object detection tasks \cite{he2020momentum}, sometimes even outperforming supervised learning. Inspired by such findings, we introduce CL into IIR with appropriate adaptations in this paper, in order to obtain discriminative and compact representations suitable for IIR. To achieve this goal, the key is to solve a {\em pretext task} by training a contrastive model, for which we propose to use {\em instance discrimination} as the pretext task. Instance discrimination learns to discriminate between individual image instances without any class labels, where each instance itself is treated as a distinct class. As such, instance discrimination based CL can learn more fine-grained features and thus would potentially be a good approach to IIR.

More specifically, we hypothesize that a CL model pre-trained for instance discrimination would learn better features (in terms of their suitability for IIR) than features learned by models pre-trained for classification. To verify this hypothesis, we first train a CL model for instance discrimination using contrastive loss \cite{wu2018unsupervised} to serve as a fixed feature extractor, and then transfer the learned features to IIR tasks. However, we conjecture that such transferred features from a CL model would be too generic to match specific tasks very well. Therefore, we propose to fine-tune our CL model with an  average precision (AP) loss. In summary, this paper makes the following contributions:

\begin{itemize}
    \item We are the first to introduce instance discrimination based contrastive learning (CL) into the field of IIR, for effective representation learning.
    \item We propose a fine-tuning strategy with an AP loss to tailor the generic feature representations learned by CL to specific IIR tasks.
    \item We empirically evaluate the proposed contrastive method as a feature extractor with and without fine-tuning for instance retrieval on the challenging Oxford and Paris datasets, and demonstrate its superior performance as compared to the off-the-shelf features learned by a popular DNN model pre-trained on ImageNet for classification.
\end{itemize}

The paper is organized as follows. Section II discusses related work, Section III describes our methods in detail, Section IV presents our experimental study, and Section V concludes this paper.

\section{Related Work}

Content Based Image Retrieval (CBIR) is a long-established research area that deals with searching databases for similar content. In general, a CBIR task belongs to one of two different groups: Category-level Image Retrieval (CIR) and Instance-level Image Retrieval (IIR) \cite{gordo2017beyond}. 
CIR aims to find all images of the same category as the query image (e.g., dogs, cars), while IIR aims to find all images that contain a particular instance that is also contained in the query image (e.g., the Eiffel Tower, BMW's logo) but may be captured under different conditions. This paper focuses on IIR tasks which are more challenging. To achieve accurate and efficient instance retrieval from a large collection of images, developing compact yet discriminative representations is at the core of IIR. Conventional instance retrieval methods relied on bag-of-visual-words encoding that builds on handcrafted local invariant features such as SIFT \cite{lowe2004distinctive} and SURF \cite{bay2008speeded}. But ever since the first time when DNNs demonstrated impressive performance on image classification \cite{krizhevsky2012imagenet}, DNN-based image representation has been wildly used for image retrieval. The most common method directly uses the neural activation in intermediate layers of DNNs to form global or local descriptors. These off-the-shelf features \cite{azizpour2015generic} have been proved to be more effective than handcrafted features in instance retrieval tasks. In addition, feature pooling or aggregation techniques also play an important role in obtaining a compact representation for instance retrieval.  Conventional encoding techniques such as VLAD \cite{gong2014multi} or Fisher vectors \cite{perronnin2015fisher} aggregate handcraft local patches to build a global descriptor, while BoW \cite{mohedano2016bags} can encode convolutional features for instance search. A more sophisticated aggregation method, R-MAC  \cite{tolias2015particular}, uses max-pooling of activations over regions of different scales to obtain compact representations. Other more sophisticated aggregation methods include SPoC \cite{babenko2015aggregating}, CroW \cite{kalantidis2016cross} and GeM \cite{radenovic2018fine} which aim to highlight feature importance or reduce the undesirable influence of bursty descriptors of some regions.

However, off-the-shelf features trained on ImageNet for classification may not be suitable for instance retrieval, which needs to differentiate between instances of the same category but the features for classification are not sensitive to that. Recently, a technique for obtaining discriminative representations emerged and is called {\em contrastive learning} (CL). Unlike supervised learning, a CL model is trained in an self-supervised manner to solve a pretext task. Among the many pretext tasks proposed, we suspect that instance discrimination might be the most suitable one for instance retrieval since it strives to discriminate among individual instances without any concept of categories. In instance discrimination \cite{wu2018unsupervised}, each instance is considered as a distinct class and a neural network is trained to maximize the agreement among many augmented views of each instance. The different views of the same image are defined as positive pairs and views from different images are negative pairs. This training process allows the CL model to gain the ability to discriminate among individual instances and thus would be a good candidate for the task of instance retrieval.  Latest works on CL such as MoCo \cite{he2020momentum}, SimCLR \cite{chen2020simple}, and BYOL \cite{grill2020bootstrap} have achieved state of the art performances on large-scale visual recognition tasks such as classification and object detection. 

Moreover, fine-tuning strategy have been studied extensively to learn better image representations for downstream tasks. DNNs pretrained on source large-scale datasets for image classification would learn the generic and global features for images. However, these deep features may not be sufficient for accurate instance retrieval, and fine-tuning on domain-specific datasets with ground-truth labels can make the pretained models perform much more accurate instance-level retrieval. Previous works have explored fine-tuning via pairwise ranking loss \cite{wan2014deep}, Siamese networks \cite{radenovic2018fine} and triplet network \cite{min2020two}. Inspired by \cite{gordo2016deep, radenovic2016cnn} which fine-tune models to learn instance-retrieval oriented representations, we fine-tune our CL model to tailor the learned features to instance retrieval with AP loss objective.

In summary, we propose a framework of how to adapt CL to IIR tasks, which consists of (1) a CL model that learns instance-discriminating representation and (2) a AP loss that tailor the representations to IIR.

\section{Proposed Method}

In this section, we describe our proposed method of CL based representation learning for instance retrieval. First, we train a CL model with instance discrimination as the pretext task. Second, we use the features learned from the CL model to perform instance retrieval via transfer learning. In this second step, we incorporate a fine-tuning strategy which tailors the learned features to IIR via re-training using AP loss.

\begin{figure*}[t]
    \centering
    \includegraphics[width=\textwidth]{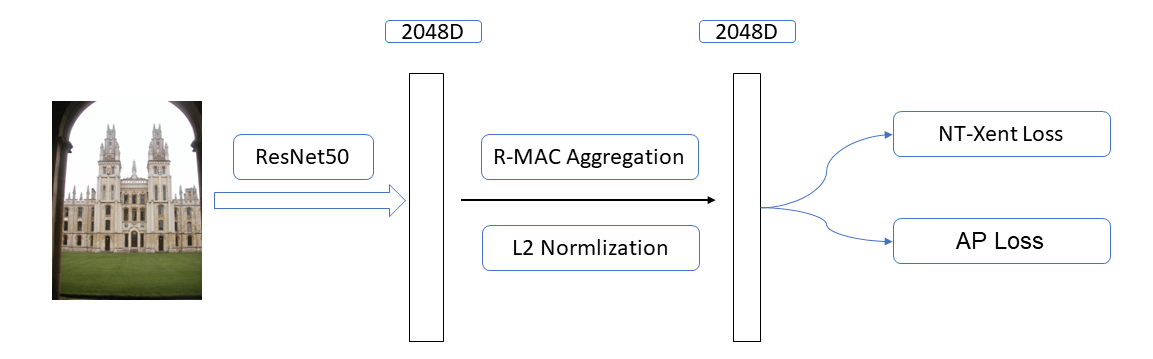}
    \caption{The training pipeline of our proposed CL method, trained with a contrastive loss and fine-tuned with an AP loss.}
    \label{fig:diagram}
    \vspace{0.1em}
\end{figure*}

\subsection{Instance Discrimination}\label{sec:ID}
Motivated by current leading contrastive methods, we choose instance discrimination as the pretext task for CL. In this task, a CL model is trained to discriminate between individual instances without any notion of semantic categories, and thus the obtained representation captures the apparent distinctions among instances. On the other hand, invariant representations are encoded via low-level image transformations (i.e., heavy data augmentation such as cropping, scaling and color jitter). With such low-level induced invariances, strong generalization to downstream tasks is achieved. Two important training approaches to the instance discrimination task are MoCo \cite{he2020momentum} and SimCLR \cite{chen2020simple}. MoCo addresses the scalability issue of storing representations of an entire database in a memory bank \cite{wu2018unsupervised}, by proposing a dynamic dictionary with a queue and a moving-averaged encoder, where samples in the queue are progressively replaced. SimCLR is a simplified CL framework that does not require any specialized architecture or memory bank, while providing better performance. Thus we adopt SimCLR to train a CL model as follows. First, we apply stochastic data augmentations on an input $\mathbf{x}$ to get two views, $\mathbf{x}_{i}$ and $\mathbf{x}_{j}$. Then, we pass the two augmented inputs sequentially through a encoder $\mathbf{f(.)}$, followed by a projection head $\mathbf{h(.)}$ that projects the output of $\mathbf{f(.)}$ to a latent space to obtain $\boldsymbol{z}_{i}$ and $\boldsymbol{z}_{j}$. The neural network (encoder and projection head) is trained with a contrastive loss function called NT-Xent loss, defined as

\begin{equation}
\ell(i, j)=-\log \frac{\exp \left(\operatorname{sim}\left(\boldsymbol{z}_{i}, \boldsymbol{z}_{j}\right) / \tau\right)}{\sum_{k=1}^{2 N} \mathbb{1}_{[k \neq i]} \exp \left(\operatorname{sim}\left(\boldsymbol{z}_{i}, \boldsymbol{z}_{k}\right) / \tau\right)}
\end{equation}
where $\mathbb{1}_{[k \neq i]} \in\{0,1\}$ is an indicator function evaluating to 1 iff $k \neq i$, $N$ is the size of each mini-batch and $\tau$ is a temperature parameter. The overall loss of each mini-batch is computed across all positive pairs in a batch as follows:

\begin{equation}
\mathcal{L}_{ID}=\frac{1}{2 N} \sum_{k=1}^{N}[\ell(2 k-1,2 k)+\ell(2 k, 2 k-1)]
\end{equation}

\begin{table*}[ht!]
\begin{center}
\scalebox{1.2}{
\begin{tabular}{|c|cccc|}
\hline
 & Oxford5k & Oxford105k & Paris6k & Paris106k \\
\hline\hline
Baseline & 65.4 & 58.9 & 77.8 & 66.7 \\
Contrastive Learning & 68.2 & 60.6 & 78.9 & 69.3 \\
CL (finetued) & \textbf{71.4} & \textbf{65.1} & \textbf{81.6} & \textbf{71.9}\\
\hline
\end{tabular}
}
\caption{Performance evaluation using mAP (\%) on Oxford and Paris datasets.}
\label{tab:ensemble_attack}
\end{center}
\end{table*}

\subsection{Fine-tuning with AP Loss}

While earlier work shows that features extracted from off-the-shelf pre-trained DNNs are more effective than handcrafted features, it was later found that such DNNs could be trained or fine-tuned specifically for the instance retrieval task, to optimize the similarity ranking instead of classification, for example by leveraging other loss functions \cite{cakir2019deep, revaud2019learning}. Thus we propose to fine-tune the above CL model, as obtained from Section~\ref{sec:ID}, using an augmented loss that optimizes for higher average precision (AP).

Let $\mathcal{B}=\left\{I_{1}, \ldots, I_{B}\right\}$ denote a batch of images and $F=\left\{f_{1}, \ldots, f_{B}\right\}$ are their corresponding features from the last convolutional layer. During each training iteration, we compute the  $\mathrm{mAP}_{B}$ over the batch. We consider each image in the batch  as a query and compare it to all the other images in the batch. The similarity score of image $I_{j}$ with respect to a query $I_{i}$ is computed as $S_{i j}=f_{i}^{\top} f_{j}$, which will determine a ranked list of images for each query $I_{i}$. Based on this list, we compute the Average Precision for each image $I_j$ in the batch, as

\begin{align}\label{eq:AP}
\mathrm{AP}_j &=\sum_{i=1}^{B} \operatorname{Prec}(i) \triangle \operatorname{Rec}(i) \\
&=\sum_{i=1}^{B} \operatorname{Prec}(i)(\operatorname{Rec}(i)-\operatorname{Rec}(i-1)) .
\end{align}
where $\operatorname{Prec}(i)$ and $\operatorname{Rec}(i)$ are the precision and recall evaluated at the $i$-th position in the ranked list, respectively. Precision is the ratio of retrieved positive images to the total number retrieved and recall is defined as the ratio of the number of retrieved positive images to the total number of positive images in the dataset. 

Then, we compute mAP for this batch by averaging over all the images in the batch as $\operatorname{mAP}_{\mathrm{B}}=\frac{1}{B} \sum_{j=1}^{B} \mathrm{AP}_{j}$. Since we want to maximize mAP on the training set, the AP loss is thus defined as 

\begin{equation}\label{eq:APloss}
\mathcal{L}_{AP}=1-\operatorname{mAP}_{\mathrm{B}}
\end{equation}

AP loss is widely used in previous works and there are many studies that address the non-differentiability in optimizing AP loss \cite{oh2016deep, rolinek2020optimizing}. The methods of optimizing $\mathcal{L}_{AP}$ for IIR  are also making rapid progress \cite{cakir2019deep, revaud2019learning, brown2020smooth} and we adopt the approach from \cite{revaud2019learning} in this paper. The training pipeline of our method is shown as Fig. \ref{fig:diagram}.

\begin{figure*}[t]
\centering 
    \begin{minipage}[b]{\textwidth} 
        \begin{subfigure}{.24\textwidth}
          \centering 
          \includegraphics[width=\linewidth]{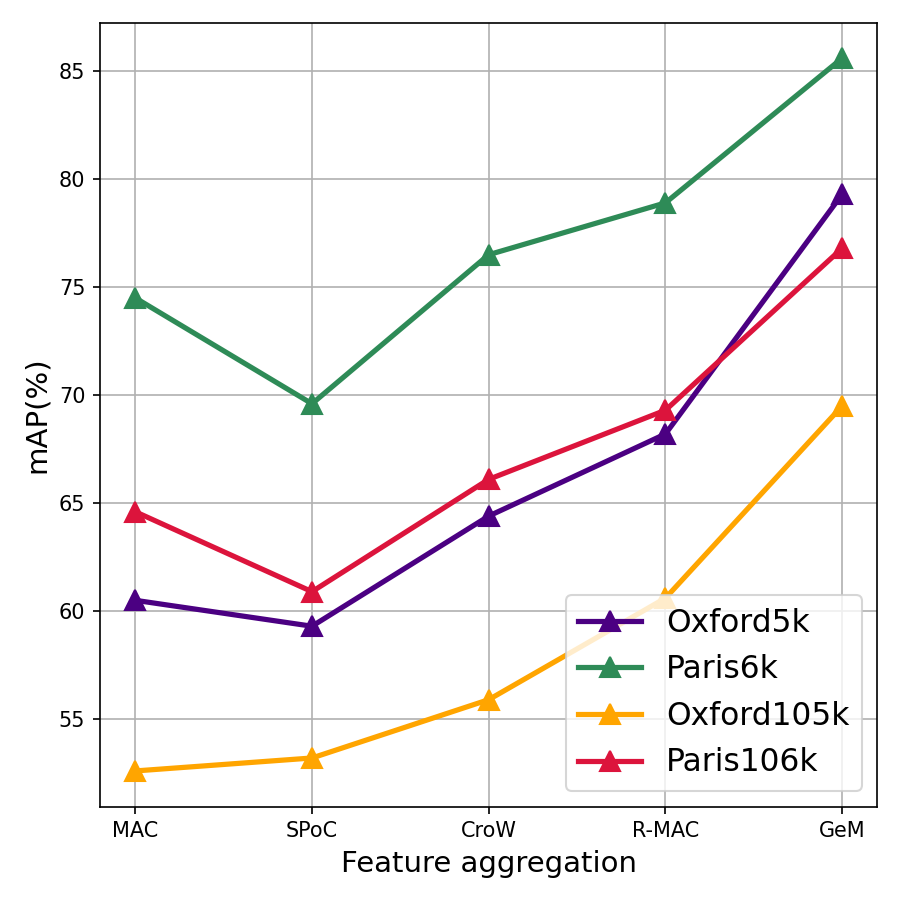}
          \caption{Impact of feature aggregation methods}
        \end{subfigure}
        \hspace{.1em}
        \begin{subfigure}{.24\textwidth} 
          \centering           \includegraphics[width=\linewidth]{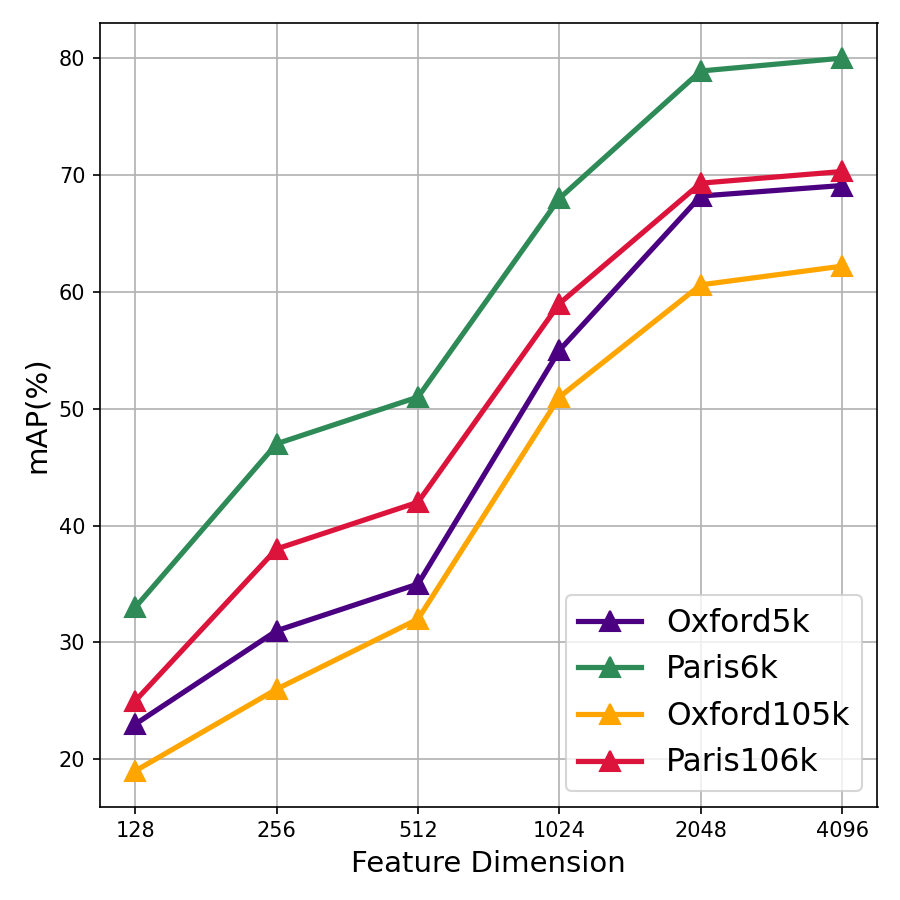}
          \caption{Impact of global feature dimension}
        \end{subfigure}%
        \hspace{.1em}
        \begin{subfigure}{.24\textwidth}
          \centering 
          \includegraphics[width=\linewidth]{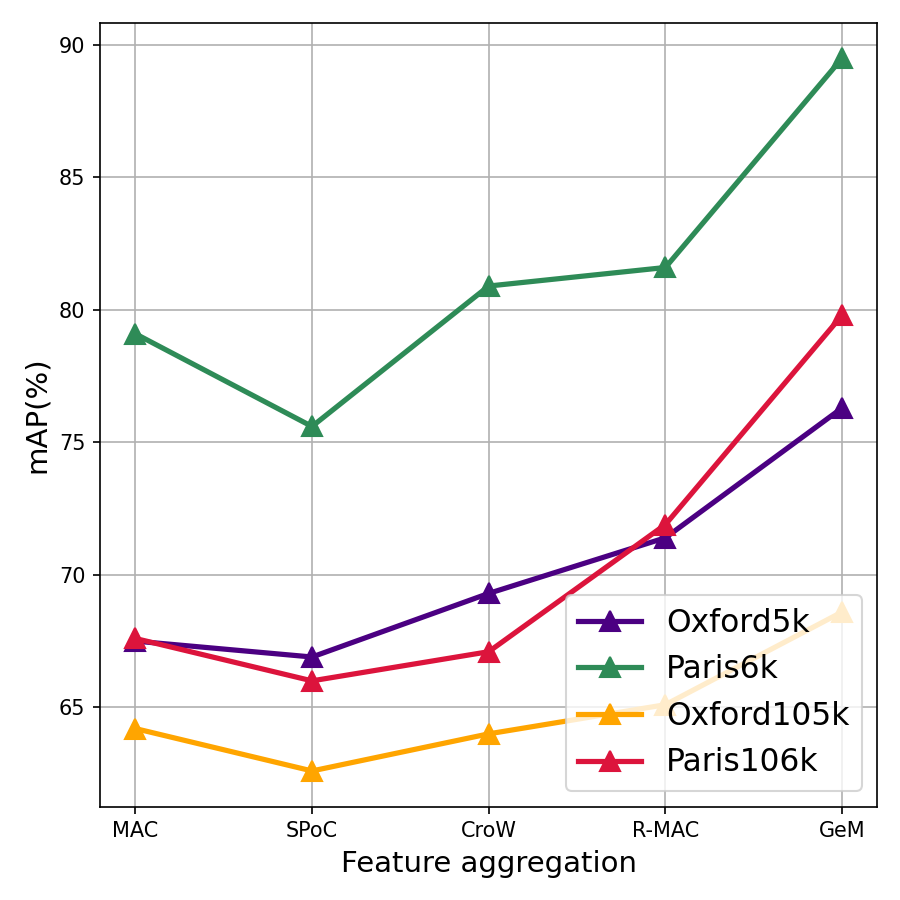}
          \caption{Impact of feature aggregation methods}
        \end{subfigure}
        \hspace{.1em}
        \begin{subfigure}{.24\textwidth} 
          \centering           \includegraphics[width=\linewidth]{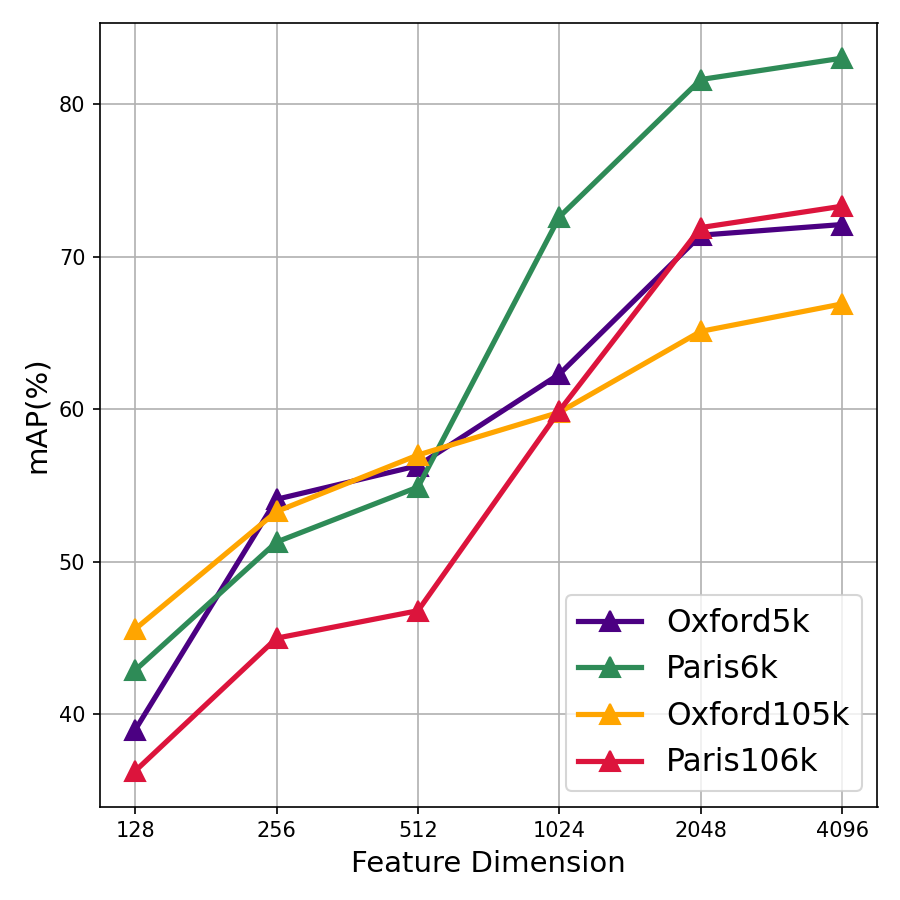}
          \caption{Impact of global feature dimension}
        \end{subfigure}%
    \caption{(a) and (b) are ablation studies on \textbf{pretrained} CL, (c) and (d) are ablation studies on \textbf{fine-tuned} CL.}
    \label{fig:sensitivity}
    \end{minipage}
\end{figure*}

\section{Experiments}

\subsection{Experiment setup}
\textbf{Datasets.} We evaluate models on two standard benchmark datasets: Oxford Buildings dataset\cite{philbin2007object} and Paris Buildings dataset \cite{philbin2008lost}. The details of the two datasets are described below:
\begin{itemize}
\item Oxford5k and Oxford105k: Oxford5k contains 5,062 images for 11 Oxford buildings. Each building comes with 5 queries---five images each manually annotated by a bounding box indicating the query instance. This gives a set of 55 queries over which an instance retrieval system can be evaluated (ground truth labels are provided for the 5,062 images). In addition, another set of 100,000 ``distractor'' images are added to obtain Oxford105k. We train and test on both datasets.

\item Paris6k and Paris106k: Paris6k contains 6,412 images collected from Flickr about Paris landmarks. Among them, 12 which have associated bounding box annotations and correspond to 12 different Paris landmarks serve as queries. The structure and labels are in the same style as Oxford5k. Similarly, an additional set of 100,000 distractor images are also added to obtain the Paris106k dataset.
\end{itemize}

\textbf{Evaluation metric.} We use the mean Average Precision (mAP) to measure the retrieval performance. The Average precision (AP) is first computed as the area under the precision-recall curve for a query, and then the mAP is computed by averaging the AP over all the query images. 

\textbf{Baseline.} Since the purpose of this work is to explore better DNN-based representation for instance retrieval, thus we use a standard ResNet50 \cite{he2016deep} trained on ImageNet for classification as a fixed feature extractor as our baseline model.

\textbf{Training details.} Training images are cropped to a fixed size of 800 × 800 (note that during test we feed the original images to the network to obtain the feature vectors). For data augmentation we use random crop and resize (with random flip), color distortions, and Gaussian blur. We use ResNet-50 as the base backbone, optimized using Adam optimizer with learning rate of 10e-3 and weight decay of 10e-6. The network is trained with batch size 512 for 50 epochs.

\textbf{Retrieval process.} Once the training is finished, the query regions are cropped and then used as input to the network. Testing images from datasets are also fed to the trained network. Since the output from the last convolutional layer preserve more structural details and hence is especially beneficial for instance retrieval, we use that as our features which has been proved effective \cite{razavian2016visual}. After we obtain the feature maps of size $\mathbb{R}^{ C\times H \times W}$, we use R-MAC \cite{tolias2015particular} over the spatial dimension to obtain compact and fixed representations of size $\mathbb{R}^C$. The final feature vector size is $C=2048$. The extracted feature vectors are post-processed by applying $L_{2}$ normalization, PCA-whitening, and $L_{2}$ normalization again. Image retrieval is simply performed by exhaustive searching over the entire database of feature vectors for each query feature vector, using the cosine similarity which is equivalent to the inner product of two normalized vectors. The similarity scores allow us to find the top-matching images to the query. 

\begin{figure*}[ht!]
\centering 
\includegraphics[width=0.9\linewidth]{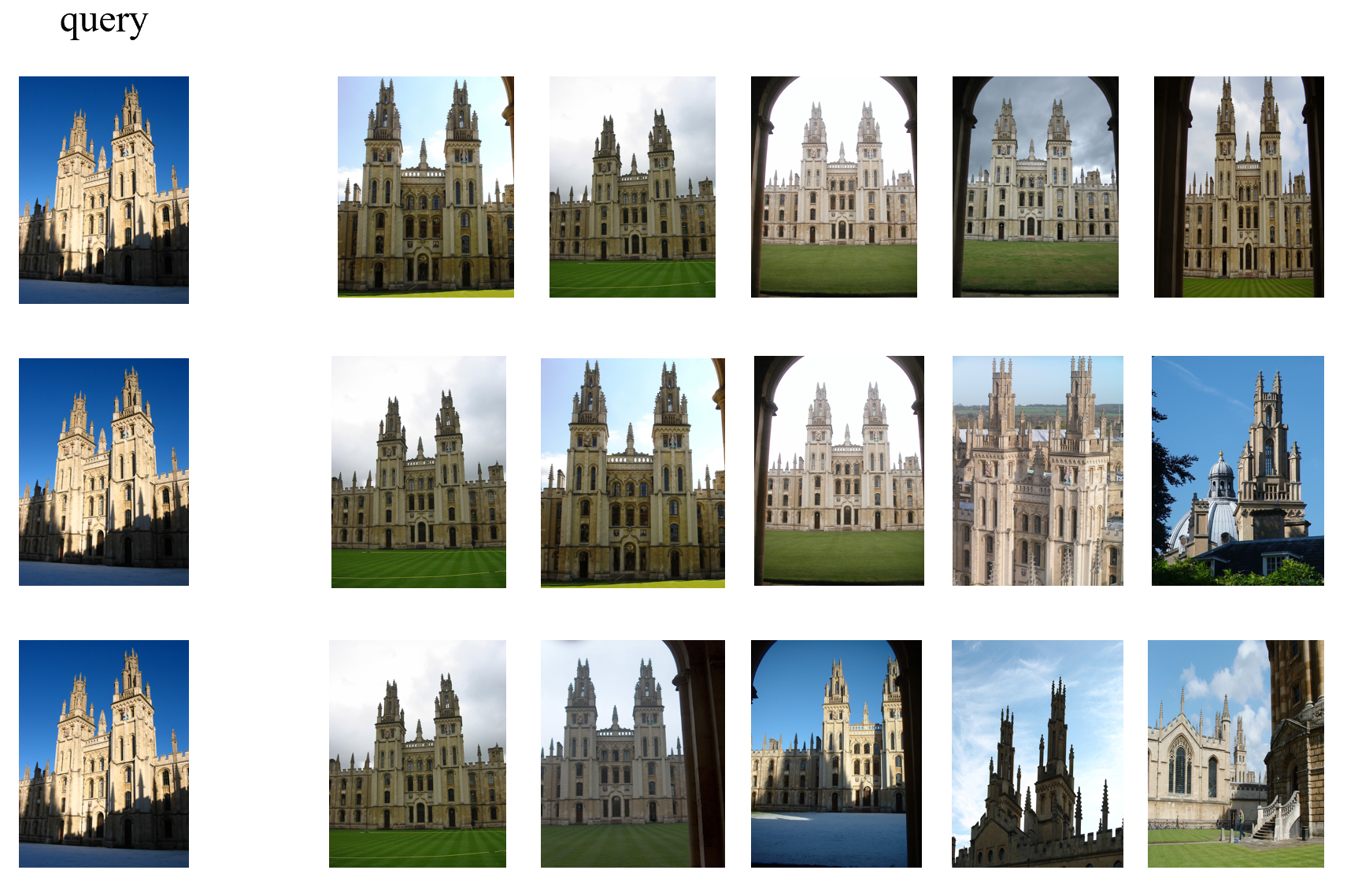}
\vspace{-0.3em}
\caption{The top 5 retrieved images by different methods: \textbf{Finetuned CL} (top) vs. \textbf{pretrained CL} (middle) vs. \textbf{pretrained ResNet-50} (bottom).}
\label{fig:viz}
\hspace{.3cm}
\end{figure*}

\subsection{Experimental results}

We now compare the results obtained by our CL model with the baseline. Note that our goal is not to beat the state of the art, rather, we aim to verify our hypothesis that the features from contrastive learning model are more suitable for instance retrieval than the features from a standard pretrained model for classification. The results for the pretained and fine-tuned CL model and baseline are summarized in TABLE I. The first set of experiments evaluate CL as a fixed feature extractor to obtain representation, where we can see that it consistently outperforms the baseline on all datasets. Moreover, we can see that the CL model fine-tuned with AP loss could boost the performance by a large margin; for instance, the mAP increases from 60.6\% to 65.1\% on the Oxford105k dataset. The results in TABLE I clearly indicates that features obtained by CL model, whether pretrained or fintuned, capture more distinctness of every instance and are more discriminative, hence are more suitable for instance retrieval tasks. 

In addition, we further conduct qualitative analysis to visually demonstrate how CL based methods can learn better representations for instance retrieval. As we can see from Figure \ref{fig:viz}, using features from pretrained and fine-tuned CL models can retrieve more correct images than the pretrained ResNet-50 classifier.

\subsection{Ablation study}
In this section, we conduct a series of ablation studies to identify the influence of some factors on the performance evaluation, namely the effects of feature aggregation methods and global feature dimension.

\textbf{Feature aggregation methods.} The methods of aggregating convolutional feature maps into a compact feature vector plays an important role in the final features.  We consider five methods for comparison: MAC \cite{razavian2016visual}, R-MAC \cite{tolias2015particular}, SPoC \cite{babenko2015aggregating}, CroW \cite{kalantidis2016cross} and GeM \cite{radenovic2018fine}. These aggregation methods aim to highlight feature importance or reduce the undesirable influence of noisy regions. We evaluate both the pretrained and fine-tuned CL models on the Oxford and Paris datasets. The results are reported in Figure 2 (a) and Figure 2 (c). We observe that the different ways to aggregate the same features from CL lead to big differences in the retrieval performance. Among the five feature aggregation methods, GeM \cite{radenovic2018fine} works the best and outperforms the others by a large margin, over 10\% on all datasets.

\textbf{Global feature dimension.} 
We vary the size of global feature dimension from 128 to 4096, and evaluate both the pretrained and fine-tuned CL models. The results are reported in Figure 2 (b) and Figure 2 (d), where it can be observed that higher dimension features obtain better performance. The possible reason is that higher dimension features usually capture more underlying semantics of an instance and thus are more helpful for retrieval. On the other hand, this advantage tends to saturate when the dimension is larger than 2048, which is likely because the features start to become redundant when the dimension becomes very large.

\section{Conclusion and Future work}

In this paper, we introduce contrastive learning (CL) to the task of instance retrieval. We also propose a find-tuning strategy to tailor the generic representation learned by pre-trained CL. We empirically show that the representation derived from both pre-trained and fine-tuned CL models  outperform off-the-shelf features from a popular pre-trained DNN for classification by large margins, and the fine-tuned CL model performs the best. In future work, we would like to design more effective fine-tuning strategies for CL models and tackle more challenging instance retrieval datasets.

\bibliographystyle{IEEEtran}
\bibliography{ref.bib}

\end{document}